\documentclass{article}

\PassOptionsToPackage{numbers, compress}{natbib}

\usepackage[final]{neurips_2024}
\usepackage{amsmath}
\usepackage{float}
\usepackage{graphicx}
\usepackage{adjustbox}

\usepackage[utf8]{inputenc} 
\usepackage[T1]{fontenc}    
\usepackage{hyperref}       
\usepackage{url}            
\usepackage{booktabs}       
\usepackage{amsfonts}       
\usepackage{nicefrac}       
\usepackage{microtype}      
\usepackage{xcolor}         

\title{Improving LLM Reasoning with Multi-Agent Tree-of-Thought Validator Agent}

\author{%
  Fatemeh Haji \\
  Secure AI and Autonomy Lab\\
  University of Texas at San Antonio\\
  \texttt{fatemeh.haji@utsa.edu} \\
  \And
  Mazal Bethany \\
  Secure AI and Autonomy Lab\\
  University of Texas at San Antonio\\
  \texttt{mazal.bethany@utsa.edu} \\
  \And
  Maryam Tabar \\
  University of Texas at San Antonio\\
  \texttt{maryam.tabar@utsa.edu} \\
  \And
  Cho-Yu Jason Chiang \\
  Peraton Labs\\
  \texttt{jchiang@peratonlabs.com} \\
  \And
  Anthony Rios \\
  University of Texas at San Antonio\\
  \texttt{anthony.rios@utsa.edu} \\
  \And
  Peyman Najafirad \\
  Secure AI and Autonomy Lab\\
  University of Texas at San Antonio\\
  \texttt{peyman.najafirad@utsa.edu}
}

\begin{document}

\maketitle

\begin{abstract}

Multi-agent strategies have emerged as a promising approach to enhance the reasoning abilities of Large Language Models (LLMs) by assigning specialized roles in the problem-solving process. Concurrently, Tree of Thoughts (ToT) methods have shown potential in improving reasoning for complex question-answering tasks by exploring diverse reasoning paths.
A critical limitation in multi-agent reasoning is the 'Reasoner' agent's shallow exploration of reasoning paths. While ToT strategies could help mitigate this problem, they may generate flawed reasoning branches, which could harm the trustworthiness of the final answer.
To leverage the strengths of both multi-agent reasoning and ToT strategies, we introduce a novel approach combining ToT-based Reasoner agents with a Thought Validator agent. Multiple Reasoner agents operate in parallel, employing ToT to explore diverse reasoning paths. The Thought Validator then scrutinizes these paths, considering a Reasoner's conclusion only if its reasoning is valid. This method enables a more robust voting strategy by discarding faulty reasoning paths, enhancing the system's ability to tackle tasks requiring systematic and trustworthy reasoning.
Our method demonstrates superior performance compared to existing techniques when evaluated on the GSM8K dataset, outperforming the standard ToT strategy by an average 5.6\% across four LLMs.
The code and related content can be found in: \url{https://github.com/SecureAIAutonomyLab/MA-ToT}

\end{abstract}

\section{Introduction}

LLMs have demonstrated remarkable capabilities across various tasks, yet they often struggle with complex reasoning, particularly in situations where human-like reasoning capabilities are crucial~\cite{zelikman2023parsel}. To address this, multi-agent strategies have emerged as a promising method to enhance LLM reasoning. Using multi-agent strategies, multiple specialized agents collaborate, with each agent assigned distinct roles in the problem-solving process. By allowing different agents to tackle various aspects of a task, we they are able to utilize specialized expertise to each phase of the task to improve performance~\cite{ijcai2024p890}. This has been shown to improve the quality of answers in reasoning-intensive domains. However, despite the promise of multi-agent reasoning, one critical limitation remains: Reasoner agents often explore reasoning paths shallowly, failing to fully consider the complexity of the problem space. Tree of Thoughts (ToT) methods offer a potential solution to this limitation by encouraging a more systematic exploration of multiple reasoning paths. ToT allows LLMs to simulate human-like thought processes by branching out and examining various possibilities before converging on a solution~\cite{yao2024tree}. By enabling LLMs to consider diverse reasoning pathways, ToT can mitigate the shallow exploration issue seen in some other multi-agent systems. However, while ToT encourages exploration, it also introduces a new challenge: the risk of generating flawed reasoning branches. Without proper validation, these erroneous paths could lower the overall trustworthiness of the final answer.

To address these challenges, we propose a novel approach that combines the strengths of multi-agent reasoning with ToT while introducing a critical validation mechanism. In our framework, multiple Reasoner agents operate in parallel, each employing ToT to explore different reasoning paths. These Reasoner agents are supported by a Thought Validator agent, which evaluates the proposed reasoning branches produced by the Reasoners. The Validator discards faulty reasoning branches, ensuring that only logically sound paths contribute to the final decision. This approach allows for both exploration of the problem space and increased reliability of the answers by eliminating flawed reasoning paths before they can impact the outcome. Our proposed approach is evaluated on the GSM8K dataset \cite{cobbe_training_2021}, a benchmark known for its challenging arithmetic reasoning tasks. Results show that our method outperforms existing techniques, demonstrating improved accuracy and trustworthiness in solving complex reasoning problems.

Our key contributions are as follows:
\begin{itemize}
\item The integration of ToT into a multi-agent reasoning framework.
\item The introduction of a novel Thought Validator agent that evaluates and filters reasoning branches produced by Reasoner agents.
\item Experimental results on the GSM8K dataset demonstrating improved accuracy and performance in complex arithmetic reasoning tasks compared to existing techniques.
\end{itemize}



\section{Background} \label{sec:background}

\subsection{Multi-agent Systems for Enhancing LLM Reasoning}

Recent work has demonstrated various approaches to enhance LLM outputs. Over-generation and reranking strategies have shown promising results, with RANKGEN~\cite{krishna2022rankgen} introducing a ranking model that scores generated outputs based on relevance and coherence, and Faithfulness-Aware Decoding~\cite{wan2023faithfulness} demonstrating how different decoding strategies affect output quality. While these methods are effective for general text generation, complex reasoning tasks often benefit from more structured approaches. Multi-agent systems have emerged as a particularly effective framework, improving performance on reasoning-based tasks by distributing tasks among specialized agents~\cite{du_improving_2023, talebirad_multi-agent_2023}. For example, CausalGPT~\cite{tang_towards_2023} introduces evaluative layers to verify the reasoning branches produced by LLMs, while the Counterfactual Multi-Agent Debate (CFMAD) framework~\cite{fang_counterfactual_2024} provides an innovative method to mitigate the potentially biased reasoning branches of LLMs by assigning agents to fixed roles to generate justifications from specific perspectives, and a third-party judge evaluates these arguments to decide the most rational outcome. 
Despite these advancements, current methods still suffer from shallow sampling of reasoning paths or majority vote schemes. These techniques can overlook critical inferential errors and are especially prone to early-stage errors, which can propagate through multiple rounds of reasoning. This limitation is especially problematic in complex scenarios where systematically evaluating and eliminating incorrect options is crucial. Recent research has demonstrated that LLMs can effectively identify both factual and inferential mistakes~\cite{li_halueval_2023}, making the integration of a dedicated verification component in multi-agent systems particularly beneficial for assessing the faithfulness and reliability of generated solutions.

\subsection{The Role of the 'Reasoner' Agent in Multi-Agent Frameworks}

Within multi-agent architectures, the Reasoner agent plays a pivotal role. It serves as the system’s core decision-maker, ensuring that valid conclusions are derived from the reasoning process. However, Reasoner agents in current frameworks often struggle to systematically evaluate and eliminate incorrect reasoning paths, particularly in more challenging problem spaces. This bottleneck highlights the need for more advanced reasoning strategies to be integrated into the Reasoner agent. CFMAD has also shown that checking all available options can enhance the overall ability of the multi-agent systems~\cite{fang_counterfactual_2024}.

\section{Method} \label{sec:method}

\begin{figure*}[t]
\centering
\includegraphics[width=\textwidth]{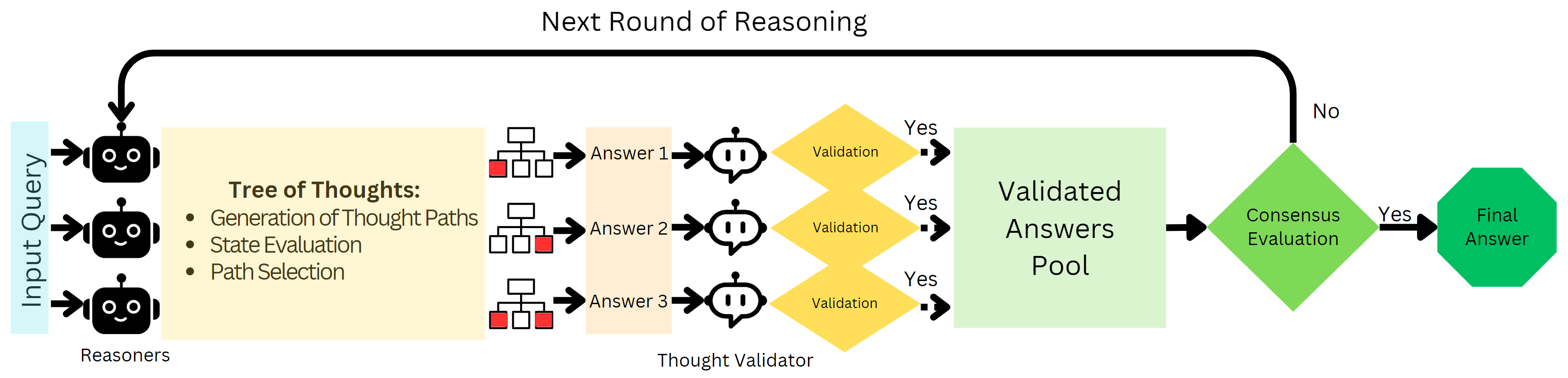}
\caption{The process starts with a query being processed by multiple Reasoner agents. Each Reasoner agent explores various reasoning paths using the ToT strategy, which includes decomposition of thought steps, generation of paths, state evaluation, and path selection. The Thought Validator agent then evaluates the proposed reasoning branches, followed by a consensus-based voting mechanism. If consensus is not reached, a new reasoning round is initiated with feedback incorporation.}
\label{fig:idea}
\end{figure*}

We propose a novel multi-agent reasoning framework that integrates the ToT strategy with a robust validation mechanism to enhance complex problem-solving. Our approach employs multiple concurrent Reasoner agents, each using ToT to explore diverse reasoning paths. At each tree level, a state evaluation agent scores the generated reasoning, with the highest-scored reasoning expanded in the subsequent level. Upon reaching the final tree level, each Reasoner agent produces a proposed reasoning chain composed of the chain of the highest-scored reasoning in the tree. These reasoning branches are then independently assessed by a Thought Validator agent to either validate or invalidate the proposed reasoning. We then use a consensus-based voting mechanism, where verified reasoning paths contribute to the vote, and invalidated ones are abstained. If consensus is not reached, we initiate a new reasoning round, incorporating feedback from the Thought Validator on the reasoning branch to refine the next reasoning round. Our proposed framework is illustrated in Figure \ref{fig:idea}.
\subsection{Reasoner Agent}
The Reasoner agents in our multi-agent architecture employ the ToT strategy, which enables structured exploration of reasoning paths in parallel. ToT improves upon Chain of Thought (CoT) prompting~\cite{wei_chain--thought_2023} by enabling parallel exploration and dynamic path evaluation. While CoT follows a single, linear path, ToT actively explores and evaluates multiple reasoning paths, making it better suited for complex problems that benefit from diverse thought exploration~\cite{yao_tree_2023}.
We formalize the reasoning process as a search over a tree of states. Let $Q$ denote the input prompt or query, and each Reasoner agent $R_i$ constructs a Tree of Thoughts $T_i(Q)$, where each node represents a state $s_t$. A state $s_t = [Q, z_1, z_2, \dots, z_t]$ consists of the problem $Q$ and a sequence of intermediate reasoning steps up to that point $z_1, z_2, \dots, z_t$, with each step $z_j$ being a coherent unit of reasoning generated by the language model.


\textbf{Step 1: Decomposition and Generation of Thought Paths}

The process is decomposed into intermediate thought steps using LLM prompting. For each state $s_t$, the next potential thought $z_{t+1}$ is generated by the \textit{Thought Generator} $G(p_\theta, s_t, k)$, where $p_\theta$ denotes the language model. The Reasoner agents explore multiple branches from any given state $s_t$, corresponding to different continuations of the reasoning process. This approach ensures that the exploration process covers a diverse range of possible solutions, avoiding the linearity of CoT and allowing reconsideration of earlier steps.

\newpage
\textbf{Step 2: State Evaluation and Path Selection}

To evaluate each state $s_t$, we introduce a state evaluation agent that assigns a score to the generated reasoning. This evaluation can be implemented through prompting, where the state evaluation agent assesses the quality and potential of each reasoning step.
At each tree level, the highest-scored reasoning is selected for expansion in the subsequent level. This process continues until the final tree level is reached. The selection mechanism can be formalized as:
\begin{equation*}
s_{t+1}^* = \arg\max_{s_{t+1}} V(p_\theta, s_{t+1})
\end{equation*}
where $V(p_\theta, s_{t+1})$ is the evaluation score assigned by the state evaluation agent.

\textbf{Step 3: Reasoning Branch Construction}

Upon reaching the final tree level, each Reasoner agent constructs a proposed reasoning chain. This chain is composed of the highest-scored reasoning steps from each level of the tree. Formally, the reasoning branch $C_i$ for Reasoner Agent $R_i$ can be represented as:
\begin{equation*}
C_i = [z_1^, z_2^, \dots, z_T^*]
\end{equation*}
where $z_t^*$ is the highest-scored reasoning step at level $t$ of the tree.
\subsection{Thought Validator Agent}
The Thought Validator agent, inspired by the role of a teacher providing feedback to students, plays a crucial role in assessing the validity of the reasoning branches produced by the Reasoner agents. Much like a teacher helping students refine their answers, this agent independently evaluates each proposed reasoning branch to either validate or invalidate it. For each reasoning branch $C_i$, the Thought Validator agent performs several key steps. It begins with a logical consistency check to evaluate the internal logic and coherence of the reasoning chain, similar to how a teacher might assess a student's argument. This is followed by a factual accuracy assessment to verify any factual claims made within the reasoning, akin to a teacher fact-checking a student's work. Finally, the agent conducts a completeness evaluation to ensure that the reasoning branch adequately addresses all aspects of the original query, much as a teacher would ensure a student's response fully answers the question. Through this comprehensive process, the Thought Validator agent ensures the robustness and reliability of the reasoning branches, ultimately helping to improve the quality of the final output.
Based on these assessments, the Thought Validator assigns a binary validation status $V_i$ to each reasoning chain:
\begin{equation*}
V_i = \begin{cases}
1 & \text{if } C_i \text{ is validated} \\
0 & \text{if } C_i \text{ is invalidated}
\end{cases}
\end{equation*}

\textbf{Consensus-Based Voting Mechanism}

After the validation process, we employ a consensus-based voting mechanism to determine the final outcome. Only validated reasoning branches contribute to the vote, while invalidated ones are abstained.
The consensus solution $S^*$ can be represented as:
\begin{equation*}
S^* = \arg\max_S \sum_{i=1}^{N} V_i \cdot \delta(S = S_i)
\end{equation*}
Where $S_i$ represents the solution derived from reasoning branch $C_i$, $V_i$ is the validation status of $C_i$, $\delta$ is an indicator function that returns 1 if the solutions match and 0 otherwise, and $N$ is the total number of Reasoner agents.
\subsection{Iterative Refinement}
If consensus is not reached (i.e., no solution receives a majority of validated votes), we initiate a new reasoning round. This refinement process incorporates feedback from the Thought Validator on the reasoning branches to guide the next iteration. This iterative process continues until consensus is reached or a predefined maximum number of iterations is exceeded.

\section{Experiments} \label{sec:experiments}


\textbf{Dataset:} GSM8K~\cite{cobbe_training_2021} is a dataset of 8.5K high-quality linguistically diverse grade school math word problems created by human problem writers. GSM8K is widely recognized as a benchmark for testing arithmetic reasoning in LLMs. The dataset comprises complex, multi-step mathematical word problems that challenge both the reasoning and computation capabilities of LLMs. Our experiments utilized a random subset of 500 samples from the GSM8K dataset as the test set. Following other works on LLM reasoning on the GSM8K dataset, we evaluated the performance of reasoning approaches using accuracy as the primary metric~\cite{yao2024tree}.

\textbf{Implementation Details: } Our experiments cover two versions of OpenAI's GPT models and two versions of Meta's Llama 3.1 models~\cite{dubey2024llama}. Specifically, we use GPT-3.5-turbo-0125~\cite{openai_gpt35turbo_2024} and GPT-4o-mini-2024-07-18~\cite{anonymous_gpt4omini_2024} from OpenAI, accessed through their API. For the Llama 3.1 models, we employ the 8B~\cite{Nous-Meta-Llama-3.1-8B-Instruct} and 70B~\cite{Nous-Meta-Llama-3.1-70B-Instruct} parameter variants. These models offer a range of capabilities and sizes, allowing us to explore different trade-offs between model complexity and performance in our experiments. We conduct all of our experiments on four Nvidia DGX A100 80 GB GPUs, and running all these experiments in parallel took about 18 hours. For our baseline comparisons, we employed several prompting strategies. We began with input-output (IO) prompting, a standard approach that transforms a problem input into an output by conditioning on task instructions. We then implemented more advanced techniques, including Chain of Thought (CoT)~\cite{wei_chain--thought_2023} and the original ToT strategy~\cite{yao2024tree}. For the ToT implementation, we followed the parameters used by ~\citet{yao2024tree} on the GSM8K dataset, setting a tree depth of 2 and a width of 5.
To ensure consistency across our baseline models, we used a temperature of 1 and a top\_p value of 1 for IO, CoT, and ToT. However, for our novel Thought Validator Agent, we adjusted these parameters to a temperature of 0.5 and a top\_p of 0.4. This adjustment was made to increase the determinism of the Thought Validator's outputs, as its role is to validate existing reasoning rather than generate creative solutions.

\textbf{Experimental Results: }
Table \ref{tab:results} shows the performance comparison of these different methods. Our results show that our proposed multi-agent ToT reasoner with a Thought Validator agent outperforms the other reasoning methods, showing an improvement of 8.8 percentage points over ToT for GPT-3.5-turbo (from 75.4\% to 84.2\%). We also see that while ToT and other techniques showed significant improvements over standard IO prompting when the LLM struggled with a task (such as with GPT 3.5 Turbo and Llama 3.1 8B), the performance gap narrowed considerably for problems where the model with standard IO prompting already exhibited strong capabilities (such as with GPT 4o mini, and Llama 3.1 70B). This observation suggests that the efficacy of ToT may be dependent on the complexity of the task and capability of the model, with its benefits more pronounced in challenging reasoning tasks that push the boundaries of the model's baseline abilities. The effectiveness of ToT in these scenarios can be likened to a teacher providing feedback to a struggling student, guiding them through complex problems, and reinforcing correct thought processes.



\begin{table}[t]
\centering
\resizebox{\textwidth}{!}{%
\begin{tabular}{l|c|c|c|c}
\hline
\textbf{Method} & \textbf{Gpt-3.5-turbo} & \textbf{Gpt-4o-mini} & \textbf{Llama3.1-8B} & \textbf{Llama3.1-70B} \\
\hline
Standard IO & 60.0 & 91.2 & 75.4 & 93.0 \\
CoT & 68.0 & 89.2 & 76.0 & 89.4 \\
ToT & 75.4 & 91.6 & 80.2 & 92.8 \\
MA ToT with Thought Validator & \textbf{84.2} & \textbf{92.2} & \textbf{89.0} & \textbf{94.8} \\
\hline
\end{tabular}%
}
\caption{Performance comparison of our Multi-agent ToT Reasoner with a Thought Validator compared to other LLM reasoning methods on the GSM8K reasoning dataset, evaluated across different LLMs.}
\label{tab:results}
\end{table}

\section{Limitations and Conclusion} \label{sec:limitations_and_conclusion}

While the ToT approach has shown promise in enhancing reasoning capabilities, our observations of the outputs and reasoning trees revealed several limitations that warrant further investigation. A key challenge we observed is the lack of dynamic exploration in the search space. The ToT method proposed by~\citet{yao2024tree} employs a fixed width and depth for the tree structure, which our analysis showed can lead to suboptimal performance in certain scenarios. For instance, when examining the reasoning trees for problems that could be solved efficiently without extensive reasoning, we found that the predetermined depth of exploration often introduced unnecessary complexity, potentially leading to errors or confusion in the reasoning process. Conversely, for problems requiring more in-depth analysis, we observed that the fixed depth proved insufficient, limiting the model's ability to fully explore complex reasoning paths. Additionally, our proposed approach, while addressing some of these limitations, is computationally expensive due to the use of the ToT method, which requires significant resources for generating and evaluating multiple thought paths. 
Our analysis shows that our multi-agent ToT approach with the Thought Validator requires substantially more compute resources than standard methods. In our 500-sample evaluation, the average token usage per question increases significantly: for GPT-3.5-turbo, from 256 tokens with CoT to 4000 tokens with ToT, while for GPT-4o-mini, from 341 to 10,600 tokens. Each Reasoner agent requires approximately 20 API calls per problem, with our multi-agent approach multiplying this cost by the number of agents plus validation steps.  While this increased computational investment yields meaningful improvements in reasoning accuracy—demonstrated by an 8.8 percentage point improvement for GPT-3.5-turbo—the trade-off may require careful consideration in resource-constrained environments. Furthermore, while our evaluation on GSM8K demonstrates the effectiveness of our approach for arithmetic reasoning, testing on additional reasoning-intensive benchmarks would help establish the method's generalizability across different types of reasoning tasks.

The Thought Validator agent demonstrates strong capabilities in assessing reasoning paths, and its effectiveness could be further enhanced through advanced validation techniques such as ensemble validation strategies or meta-learning approaches to improve robustness across diverse reasoning scenarios. Furthermore, while our evaluation on GSM8K demonstrates the effectiveness of our approach for mathematical reasoning, testing on additional reasoning-intensive benchmarks would help establish the method's generalizability across different types of reasoning tasks.

In conclusion, we have presented a novel approach that combines the ToT strategy with a multi-agent reasoning framework enhanced by a Thought Validator agent. Our method addresses key limitations in existing reasoning strategies for LLMs by enabling a more systematic exploration of reasoning paths while simultaneously improving the reliability of generated solutions. Experimental results on the GSM8K dataset demonstrate that our approach outperforms state-of-the-art methods, particularly for complex arithmetic reasoning tasks. Future work could explore dynamic tree structuring based on problem complexity, potentially improving efficiency and performance across a wider range of problem types.

\section{Social Impact Statement}
By improving the depth of reasoning and enabling more systematic option elimination, our approach could lead to more trustworthy AI applications. However, these advancements also raise ethical considerations regarding the deployment of highly autonomous reasoning systems, particularly in high-stakes domains. It is essential to carefully manage the use of such systems to avoid over-reliance on AI, ensuring that human oversight and accountability remain integral to decision-making processes. Additionally, the broader societal implications must be monitored to prevent unintended consequences, such as biases being amplified through algorithmic decision-making or the replacement of human expertise in fields where nuanced judgment is required.

\bibliographystyle{plainnat}
\bibliography{bibliography}

\section*{Appendix}



\subsection*{Experiment Prompts}
\label{prompts}
In our experiments, we designed a number of carefully crafted prompts to guide language models during reasoning tasks. Here are the key prompts used and their purposes:

\subsubsection*{Standard Input-Output (IO) Prompt}
The standard IO prompt is used as a baseline approach:

\begin{verbatim}
    Answer the following math problem. Your response should 
    conclude with "the answer is n", where n is a number:
    {input}
\end{verbatim}

This prompt directly asks the model to solve the math problem and provide the answer in a specific format.

\subsubsection*{Chain of Thought (CoT) Prompt}
The CoT prompt encourages the model to show its reasoning:

\begin{verbatim}
    Answer the following question: {input}
    Make a strategy, then write. Your output should be in 
    the following format:
    
    Strategy:
    Your strategy about how to answer the question.
    
    Answer:
    Your answer to the question. It should end with 
    "the answer is n", where n is a number.
\end{verbatim}

This prompt explicitly asks the model to formulate a strategy before providing an answer, leading to a more structured thought process.

\subsubsection*{Tree of Thoughts (ToT) Prompt}
Our implementation of ToT is inspired by the approach described by \citet{yao2024tree} but with specific modifications tailored to our multi-agent framework. The ToT method uses the CoT prompt as a base and applies it iteratively, allowing for branching and exploration of multiple reasoning paths. Our implementation includes the following components:

\begin{enumerate}
    \item \textbf{Thought Generation:} We use the 'sample' method for generating thoughts. This method uses the CoT prompt as a base but applies it iteratively, allowing for branching and exploration of multiple reasoning paths. The prompt flow for ToT includes:

    \begin{verbatim}
    Answer the following question: {input}
    Make a strategy, then write. Your output should be in 
    the following format:

    Strategy:
    Your strategy about how to answer the question.

    Answer:
    Your answer to the question. It should end with 
    "the answer is n", where n is a number.
    \end{verbatim}

    \item \textbf{State Evaluation:} For evaluating the generated thoughts, we employ the 'vote' method. This involves using a prompt to assign votes to different reasoning paths:

    \begin{verbatim}
    Given an instruction and several choices, decide which
    choice is most promising. Analyze each choice in detail, 
    then conclude in the last line "The best choice is {s}", 
    where s the integer id of the choice.
    \end{verbatim}

    \item \textbf{Path Selection:} We use the 'greedy' method for selecting the most promising paths to expand further. This doesn't involve a specific prompt but rather selection of the highest-scored paths from the evaluation step.
\end{enumerate}

Each step involves multiple API calls to the language model, with the generated thoughts and their evaluations guiding the exploration of the reasoning space. This approach allows for a dynamic and adaptive exploration of potential solution paths, enhancing the model's ability to tackle complex reasoning tasks.

\subsubsection*{Verifier Prompt}
A crucial component of our approach is the Thought Validator agent, which uses the following prompt:

\begin{verbatim}
    As a critical mathematical reasoning verifier, evaluate 
    the following thought process, which builds upon previous 
    steps to reach a final conclusion. Focus on:
    
    1. **Question Relevance**:
       - Ensure the entire reasoning process directly 
         addresses the original question.
       - Check if the final answer actually solves what 
         was asked.
    
    2. **Reasoning Progression**:
       - Assess logical flow and consistency, especially 
         in final steps.
       - Verify mathematical operations' correctness and 
         appropriateness.
       - Identify logical fallacies or unjustified leaps.
    
    3. **Factual Accuracy**:
       - Check accuracy and relevance of facts and numbers, 
         particularly in final calculations.
       - Spot any misuse of mathematical concepts.
    
    4. **Completeness**:
       - Ensure all necessary aspects are addressed, 
         particularly in concluding thoughts.
       - Identify significant omissions that could affect 
         the result.
    
    5. **Critical Assessment**:
       - Actively seek potential errors or weak points.
       - Don't hesitate to invalidate reasoning if 
         significant issues are found.
    
    Provide a holistic evaluation of the entire reasoning 
    process, from start to finish. Conclude with 
    "Reasoning is Valid" only if the entire process is 
    relevant, logically sound, and error-free. Otherwise, 
    conclude with "Reasoning is Invalid" and briefly 
    explain why.
\end{verbatim}

This comprehensive prompt guides the Verifier in thoroughly assessing the validity of the reasoning process, ensuring that the final answer is not only correct but also logically sound and relevant to the original question.

\subsection*{Examples}

To demonstrate the effectiveness of our approach, we show a challenging example from the GSM8K dataset using the gpt-3.5-turbo model. Using this example, we can see how the Thought Validator Agent prevents incorrect reasoning from the ToT Reasoner agents from leading to errors in the final answer.


\textbf{Problem 1:}
\textit{Last month, Tasha made \$80 from selling lemonade and mowing lawns. The first week, she mowed Kamala's lawn three times as many times as Joe's. The following week, she mowed Alba's lawn five times as Joe's. If Joe paid Tasha \$6 for her work, how much did she make from lemonade sales? \textbf{Answer: 26.}
}

We have three rounds, each involving three Reasoner agents (R1, R2, and R3). After each round, the Thought Validator Agent evaluates their reasoning.

\begin{table}[htbp]
\centering
\begin{adjustbox}{width=\textwidth,center}
\begin{tabular}{|c|p{8cm}|l|c|}
\hline
\textbf{Reasoner} & \textbf{Reasoning Summary}                & \textbf{Final Answer} & \textbf{Verified} \\ \hline
R1 & Incorrect reasoning. Algebraic error leads to the incorrect conclusion that Tasha did not mow Joe's lawn.                          & \$80                  & False            \\ \hline
R2 & Correct strategy, but incorrect total earnings from mowing lawns (\$60x instead of \$48x).                                        & \$80                  & False            \\ \hline
R3 & Accurate reasoning. Correctly calculates the total earnings from mowing lawns and finds lemonade income to be \$26.                & \$26                  & True             \\ \hline
\end{tabular}
\end{adjustbox}
\caption{(Round 1) Reasoner Outputs and Verification Status}
\end{table}

\subsection*{Round 1 Analysis}

In Round 1, Reasoner 1 (R1) incorrectly calculated that Tasha did not mow Joe’s lawn, leading to an invalid final answer of \$80. Reasoner 2 (R2) correctly identified the structure of the problem but made a calculation error, arriving at \$80. Reasoner 3 (R3) provided the correct reasoning and final answer of \$26, which was verified as valid. However we have not reached an agreement since the only valid answer is the R3 response.

{
\centering
Round 1 Conclusion: No Consensus Reached.\\
}

\begin{table}[htbp]
\centering
\begin{adjustbox}{width=\textwidth,center}
\begin{tabular}{|c|p{8cm}|l|c|}
\hline
\textbf{Reasoner} & \textbf{Reasoning Summary}                & \textbf{Final Answer} & \textbf{Verified} \\ \hline
R1 & Repeated the previous error, miscalculating Tasha's income from lawn mowing.                                                       & \$80                  & False            \\ \hline
R2 & Corrected earlier miscalculation but again used the wrong mowing total, leading to an incorrect conclusion.               & \$80                  & False            \\ \hline
R3 & Maintained the correct reasoning and final answer (\$26) as in Round 1.
    & \$26                  & True             \\ \hline
\end{tabular}
\end{adjustbox}
\caption{(Round 2) Reasoner Outputs and Verification Status}
\end{table}

\subsection*{Round 2 Analysis}

In Round 2, Reasoner 1 (R1) repeated its earlier algebraic mistake. Reasoner 2 (R2) adjusted its calculations but still produced an incorrect final answer. Reasoner 3 (R3) again provided the correct reasoning and final answer of \$26.

{
\centering
Round 2 Conclusion: No Consensus Reached.\\
}

\begin{table}[htbp]
\centering
\begin{adjustbox}{width=\textwidth,center}
\begin{tabular}{|c|p{8cm}|l|c|}
\hline
\textbf{Reasoner} & \textbf{Reasoning Summary}                & \textbf{Final Answer} & \textbf{Verified} \\ \hline
R1 & Corrects earlier algebraic error, but still does not address the lemonade sales correctly.                                            & \$80                  & False            \\ \hline
R2 & Adjusts previous mistake but there is a slight issue in the final calculation but the Validator Agent was not able to detect it.
& \$32                  & False            \\ \hline
R3 & Maintained the correct reasoning and final answer (\$26) as in Round 1.
    & \$26                  & True             \\ \hline
\end{tabular}
\end{adjustbox}
\caption{(Round3) Reasoner Outputs and Verification Status}
\end{table}

\subsection*{Round 3 Analysis}

In Round 3, Reasoner 1 corrected its earlier algebraic errors but still provided an invalid answer (\$80). Reasoner 2 finally corrected its calculations, but still has a small issue in final answer and reach to \$32. Reasoner 3 remained consistent and accurate, providing the correct answer of \$26.

{
\centering
Final Conclusion: Most frequent valid answer is \$26. Final answer: 26 \\
}

\vspace{5mm} 
\hrule 
\vspace{2mm} 

\textbf{Problem 2:}
\textit{Bob is in charge of doing laundry for a large hotel. Each room has two sheets, one comforter, twice as many pillow cases as sheets and twice as many towels as pillow cases. How many pieces of laundry are there in 80 rooms? \textbf{Answer: 26.}
}

\begin{table}[htbp]
\centering
\begin{adjustbox}{width=\textwidth,center}
\begin{tabular}{|c|p{8cm}|l|c|}
\hline
\textbf{Reasoner} & \textbf{Reasoning Summary}                & \textbf{Final Answer} & \textbf{Verified} \\ \hline
R1 & Accurate breakdown of laundry items per room and multiplication across 80 rooms. Correct total of 1200 pieces of laundry.                                                                & 1200                  & True            \\ \hline
R2 & CCorrect approach, but overestimated the number of pillowcases, leading to an incorrect total of 1280 pieces of laundry.                                                            
   & 1280                  & False            \\ \hline
R3 & Correct breakdown of laundry per room, yielding the correct total of 1200 pieces of laundry.
    & 1200                  & True             \\ \hline
\end{tabular}
\end{adjustbox}
\caption{(Round 1) Reasoner Outputs and Verification Status}
\end{table}

\subsection*{Round 1 Analysis}

In Round 1, the Thought Validator Agent evaluated the responses from all three Reasoners. Both Reasoner 1 (R1) and Reasoner 3 (R3) provided correct and consistent reasoning, each arriving at the total of 1200 pieces of laundry, which was validated as accurate. However, Reasoner 2 (R2) overestimated the number of pillowcases, leading to an incorrect answer of 1280 pieces, which was marked as invalid.

Since two verified Reasoners (R1 and R3) agreed on the correct answer, the final result of 1200 pieces of laundry was confidently returned.

{
\centering
Round 1 Conclusion: At least two verified reasoners agree. Final Answer: 1200\\
}


\end{document}